\documentclass[runningheads]{llncs}
\usepackage{graphicx}
\usepackage{amsmath}
\usepackage{amsbsy}
\usepackage{amssymb}
\usepackage{hyphenat}
\newcommand{\citep}[1]{\cite{#1}}

\begin{document}
\title{Using Subjective Logic to Estimate Uncertainty in Multi-Armed Bandit Problems}
%
\titlerunning{Subjective Logic Multi-Armed Bandits}
%
\author{Fabio Massimo Zennaro\inst{1}\orcidID{0000-0003-0195-8301} and Audun J\o sang\inst{1}\orcidID{0000-0001-6337-2264}}
%
\authorrunning{F.M Zennaro, A. J\o sang}
%
\institute{Department of Informatics, University of Oslo, 0316 Oslo, Norway
\email{ fabiomz@ifi.uio.no,josang@mn.uio.no}}
%
\maketitle              
\begin{abstract}
The multi-armed bandit problem is a classical decision\hyp{}making problem where an agent has to learn an optimal action balancing \emph{exploration} and \emph{exploitation}. Properly managing this trade-off requires a correct assessment of uncertainty; in multi-armed bandits, as in other machine learning applications, it is important to distinguish between stochasticity that is inherent to the system (\emph{aleatoric uncertainty}) and stochasticity that derives from the limited knowledge of the agent (\emph{epistemic uncertainty}). In this paper we consider the formalism of \emph{subjective logic}, a concise and expressive framework to express Dirichlet-multinomial models as subjective opinions, and we apply it to the problem of multi-armed bandits. We propose new algorithms grounded in subjective logic to tackle the multi-armed bandit problem, we compare them against classical algorithms from the literature, and we analyze the insights they provide in evaluating the dynamics of uncertainty. Our preliminary results suggest that subjective logic quantities enable useful assessment of uncertainty that may be exploited by more refined agents. 

\keywords{Multi-armed bandits \and Subjective logic \and Uncertainty.}
\end{abstract}
\section{Introduction}

The \emph{multi-armed bandit problem} is a basic instance of a reinforcement learning problem, in which an agent needs to deal with the uncertainty regarding the reward it may receive after taking an action \citep{sutton2018reinforcement,lattimore2018bandit,slivkins2019introduction}. Such an agent has to balance \emph{exploratory} actions, aimed to reduce its overall uncertainty (but potentially detrimental with respect to its objective of accumulating as much reward as possible), and \emph{exploitative} actions aimed at maximizing its rewards (but based on inherently uncertain and possibly incomplete knowledge of the value of all the actions). A multi-armed bandit problem captures the essential problem of taking actions in an uncertain context, and, as such, this model has been used to formalize challenges in many fields, from online advertisement to medical trials.

Interest in assessing and quantifying \emph{uncertainty} has increased in recent years, as a reliable estimation of uncertainty has been recognized as crucial for informed decision-making. Different frameworks to deal with uncertainty exist in the fields of logic, statistics and machine learning \citep{hullermeier2019aleatoric}.
A central challenge concerns the ability to distinguish between two main forms of uncertainty: \emph{aleatoric} uncertainty, related to the uncertainty deriving from the intrinsic stochasticity of an observed phenomenon; and \emph{epistemic} uncertainty, related to the uncertainty deriving from the limits of the inference process itself \citep{kendall2017uncertainties,hullermeier2019aleatoric}. This distinction has been studied using standard statistical models in machine learning  \cite{malinin2018predictive,kendall2017uncertainties}.

A new and promising approach to the evaluation of uncertainties is based on the use of \emph{subjective logic} \citep{jøsang2016subjective}, a formalism that allows to express specific forms of probability distributions in a simple and interpretable way. This framework has been recently used to assess uncertainties in deep networks \cite{sensoy2018evidential,kaplan2018uncertainty}.

In this paper we consider the application of subjective logic to the problem of multi-armed bandits. We propose algorithms to solve the multi-armed bandit problem based primarily on subjective logic quantities, we experimentally assess their performance alongside other classical algorithms, and we analyze the insights we may get from the estimation of uncertainty that subjective logic allows.

\section{Background}\label{sec:Background}

This section presents the basic concepts related to multi-armed bandits, uncertainty and subjective logic.

\subsection{Multi-armed Bandits}

The multi-armed bandits or $k$-armed bandit problem \citep{sutton2018reinforcement,lattimore2018bandit,slivkins2019introduction} is a learning problem in which an agent is required to learn the best action $a^{*}$ in a set of actions $\mathcal{A}$. Formally, an instance $\mathcal{B}$ of the multi-armed bandit problem is defined by a tuple:
\[
\mathcal{B}=\left\langle \mathcal{A},\mathcal{R}\right\rangle 
\]
where $\mathcal{A}=\left\{ a_{1},a_{2},...,a_{k}\right\} $ is a set
of $k$ actions, and $\mathcal{R}=\left\{ \rho_{1},\rho_{2},...,\rho_{k}\right\} $
is a set of $k$ probability distributions such that $\rho_{i}$ defines
the probability distribution over the different rewards associated with taking action $a_{i}$. Whenever the agent takes action $a_{i}$, a reward $r_{i}$ is sampled as a random variable $R_i$ from $\rho_{i}$. Few assumptions are made on the distributions $\rho_{i}$: (i) rewards are finite (i.e., $\rho_{i}$ has finite mean and variance); (ii) rewards are independent; (iii) the distribution $\rho_{i}$ is stationary. This setup resembles playing with a slot machine equipped with $k$ levers (also known as a multi-armed bandit).

We define the optimal action $a^{*}$ to be the action associated with the probability distribution $\rho^{*}$ having the highest expected value, that is,
$\rho^{*}=\rho_{i}\,\,\,s.t.\,\,\,E\left[R_{i}\right] \geq E\left[R_{j}\right]\,\,\,\forall j,1\leq j\leq k.$
Notice that the optimal action does not have to be unique. The true probability distributions $\rho_{i}$ and the expected rewards $E\left[R_{i}\right]$ are unknown to the agent. The aim of the agent is to interact with the multi-armed bandit and learn the best action $a^{*}$ as quickly as possible. To formalize this, a temporal dimension is introduced. Let $a^{(t)}$ and $r^{(t)}$ denote the action and the reward at the time-step $t$; let also $\rho^{(t)}$ and $R^{(t)}$ be the probability distribution and the random variable associated with action $a^{(t)}$. The objective of the agent is to maximize its expected rewards on a given time horizon $T$, that is:
$\max\sum_{t=0}^{T}E\left[R^{(t)}\right]$. 
The simple multi-armed bandit problem constitutes a non-trivial learning game in which the agent is required to carefully balance \emph{exploration} (trying out actions in order to learn their true rewards) and \emph{exploitation} (choosing the action estimated as the best in order to maximize the rewards).
Classical approaches to solve this problem include \emph{randomized-exploration algorithms} that relies on random actions to balance exploitation and exploration (e.g.: $\epsilon$-greedy algorithm); \emph{statistical-exploration algorithms} that estimates statistical measures to balance exploration and exploitation (e.g.:  upper-confidence bound algorithm); or, \emph{preference-based algorithms} that exploit gradient optimization to learn an optimal behavior (e.g.: gradient bandit algorithm).

\subsection{Uncertainty}

Quantifying uncertainty is a crucial challenge in order to create reliable and interpretable statistical models.
Let $\mathcal{X} = \left\{ x^{(1)}, x^{(2)}, ..., x^{(N)} \right\}$ be a set of observations generated by an unknown stochastic model $p^*(x)$. Let us define a parametric statistical model $p(x\vert \boldsymbol{\theta})$ to describe the observations $\mathcal{X}$. In order to fit the model $p(x\vert \boldsymbol{\theta})$ we learn the values of the set of parameters $\boldsymbol{\theta}$ from observations. Inevitably, this fitting procedure is affected by uncertainty, deriving both from the intrinsic randomness of the data generating process $p^*(x)$ and from the limited data to infer the parameters of the model $p(x\vert \boldsymbol{\theta})$. We can then distinguish two main forms of uncertainty:

\begin{itemize}
	\item \emph{Aleatoric uncertainty} represents the uncertainty that is inherent to the data and the process $p^*(x)$ that generated the data  \citep{kendall2017uncertainties}. No matter how much data we provide, an optimally fitted parametric model $p(x\vert \boldsymbol{\theta})$ can not have less uncertainty than the true underlying generating stochastic model $p^*(x)$; in this sense, aleatoric uncertainty can not improve beyond its limit simply providing more data \citep{kendall2017uncertainties,hullermeier2019aleatoric}.
	
	\item \emph{Epistemic uncertainty} represents the uncertainty that is due to finite sampling and the limits of inference in the model \citep{kendall2017uncertainties}. Interpolation in potentially under-sampled regions of the domain of $\mathcal{X}$ or extrapolation out-of-domain in regions uncovered by $\mathcal{X}$ is necessarily affected by uncertainty. Differently from aleatoric uncertainty, epistemic uncertainty may be improved by gathering more data, thus reducing uncertainty in-domain or out-domain. 
\end{itemize}


The present study focuses on these two forms of uncertainty\footnote{Notice that in the literature it is possible to identify additional types of uncertainties, and that the nomenclature is not universally agreed \cite{hullermeier2019aleatoric,malinin2018predictive}.}. Several measures may be adopted to estimate these forms of uncertainty \cite{malinin2018predictive}, and correctly evaluating these two forms of uncertainty would allow for a more insightful analysis of a model. For instance, high aleatoric uncertainty and low epistemic uncertainty may mean that we are dealing with an intrinsically random process that can not be characterized more precisely, even with more data; on the other hand, high epistemic uncertainty may suggest that our conclusions are based on a very limited set of observations, and that more data may be in order.

\subsection{Subjective Logic} \label{ssec:SL}

Subjective logic \cite{jøsang2016subjective} provides a concise formalism to represent Dirichlet\hyp{}multinomial and Dirichlet\hyp{}categorical models.
Given a discrete domain $\mathcal{A} = \left\{ a_1, a_2, ..., a_k \right\}$ with $k$ elements, a \emph{multinomial opinion} over this domain is a tuple:
\[ 
\omega = \left( \boldsymbol{b}, u, \boldsymbol{c} \right),
\]
where $\boldsymbol{b},\boldsymbol{c} \in \mathbb{R}_{\geq0}^k$, $u \in \mathbb{R}_{\geq0}$, subject to the constraints: $u + \sum_{i=i}^{k} b_i =1$ and $\sum_{i=1}^{k} c_i = 1$.
The vector $\boldsymbol{b}$ is interpreted as a \emph{belief vector}, expressing the degree of certainty over the $k$ elements; $u$ is an \emph{uncertainty scalar}, expressing the degree of certainty in modeling the belief; and $\boldsymbol{c}$ is interpreted as a \emph{base rate vector}, expressing a prior distribution over the $k$ elements.
Subjective modeling can also be seen as a form of hierarchical modelling, in which the belief vector captures \emph{first-order} certainty about the distribution of beliefs over the domain, and the uncertainty scalar expresses a global \emph{second-order} certainty about the modeling of beliefs.
This formulation allows for an immediate distinction of the two forms of uncertainty we are concerned with: the belief vector with its distribution over the $k$ elements expresses aleatoric uncertainty, while the uncertainty scalar expresses epistemic uncertainty.

In the subjective logic framework the probability of an element $a_i$ in the domain of definition may be computed as:
\begin{equation}\label{eq:SLprobs}
P(a_i \vert \omega) = b_i + u c_i. 
\end{equation}

Moreover, there exist mappings between an opinion $\omega = \left( \boldsymbol{b}, u, \boldsymbol{c} \right)$, an evidential Dirichlet pdf $s = \mathtt{Dir_e}\left(\mathbf{e}\right)$ with evidence parameter $\mathbf{e} \in \mathbb{R}_{\geq0}^k$, and a Dirichlet pdf $q = \mathtt{Dir}\left(\boldsymbol{\alpha}\right)$ with concentration parameter $\boldsymbol{\alpha} \in \mathbb{R}_{\geq0}^k$:
\begin{equation}\label{eq:tauW}
\tau_{W}\left(\omega\right):\begin{cases}
e_{i}=\frac{Wb_{i}}{u} & \textrm{if }u\neq0\\
e_{i}=\infty & \textrm{otherwise}
\end{cases}
\end{equation}
\begin{equation}
\sigma_{W}\left(\omega\right):\begin{cases}
\alpha_{i}=W\left(\frac{b_{i}}{u}+c_{i}\right) & \textrm{if }u\neq0\\
\alpha_{i}=\infty & \textrm{otherwise}
\end{cases}
\end{equation}
where $W$ is a non-informative prior weight normally specified equal to $2$ for consistency \cite{jøsang2016subjective}. 
Since an opinion has more degrees of parametric freedom compared to an evidential Dirichlet or a Dirichlet distribution, there is no exact inverse, and mapping back a Dirichlet distribution onto a subjective opinion requires the choice of a base rate vector $\boldsymbol{c}$:
\begin{equation}\label{eq:tauW-1}
\tau^{-1}_{W,\boldsymbol{c}}\left(\mathbf{e}\right) :\begin{cases}
b_{i}=\frac{e_i}{W + \sum_{i=1}^{k} e_i}\\
u=\frac{W}{W + \sum_{i=1}^{k} e_i}
\end{cases}
\end{equation}
\begin{equation}\label{eq:sigmaW-1}
\sigma^{-1}_{W,\boldsymbol{c}}\left(\boldsymbol{\alpha}\right) :\begin{cases}
b_{i}=\frac{\frac{\alpha_{i}}{W}-c_{i}}{1+\sum_{i=1}^{k}\left(\frac{\alpha_{i}}{W}-c_{i}\right)}\\
u=\frac{1}{1+\sum_{i=1}^{k}\left(\frac{\alpha_{i}}{W}-c_{i}\right)}
\end{cases}
\end{equation}
where, as before, we assume $W=2$ as a non-informative prior weight.

In conclusion, given an opinion $\omega = \left( \boldsymbol{b}, u, \boldsymbol{c} \right)$, it is possible to reconstruct a standard statistical Dirichlet-categorical model as:
\begin{subequations}\label{eqs:mappingSL2Ev}
\begin{align}
	\boldsymbol{\alpha} & = \sigma_{W}\left(\omega\right)\\
	q\left(\boldsymbol{\theta}\vert\boldsymbol{\alpha}\right) & = \mathtt{Dir}\left(\boldsymbol{\alpha}\right)\\
	p\left(\mathbf{a}\vert\boldsymbol{\theta}\right) & = \mathtt{Cat}\left(\boldsymbol{\theta}\right).
\end{align}
\end{subequations}

\section{Subjective logic bandit algorithms} \label{sec:SubjectiveBandit}

In this section we introduce the subjective logic bandit algorithm (SLB), an algorithm to solve the multi-armed bandit problems grounded in subjective logic. Although it is always possible to transform subjective opinions in statistical distributions using the mapping expressed in Equation \ref{eqs:mappingSL2Ev}, we explore here the possibility of working primarily in the domain of subjective logic.

\subsection{Estimating the probability of actions}

Differently from classical algorithms which may directly estimate the \emph{average reward of actions} (like $\epsilon$-greedy or UCB algorithms) or track a \emph{preference over actions} (like gradient algorithms), the central quantity considered by SLB is a multinomial opinion over $k$ available actions:
\[ 
\omega^{(t)} = \left( \boldsymbol{b}^{(t)}, u^{(t)}, \boldsymbol{c} \right),
\]
where $b_i^{(t)}$ expresses the belief in action $a_i$ being the best action at time $t$, $c_i$ is the base rate or prior belief in action $a_i$ being the best action, and $u^{(t)}$ is the global uncertainty in this estimation at time $t$. Notice that the base rate $c_i$ is not time-indexed, as it reflects common prior knowledge and it is assumed to be unchanged for the duration of the problem.

This formalism allows an agent to be initialized in a setup expressing complete ignorance by setting:
\[
\begin{cases}
b_{i}^{(0)}=0 \qquad \forall i,1\leq i\leq k\\
u^{(0)}=1\\
c_{i}=\frac{1}{k} \qquad \forall i,1\leq i\leq k.
\end{cases}
\]
This setup encodes an agent with no belief in any action being the best ($b_{i}^{(0)}=0$), no prior information to favor one action over another ($c_{i}=\frac{1}{k}$), and complete uncertainty about its own model ($u^{(0)}=1$).

In order to sample actions using standard routines we can compute a distribution of probability of actions given an opinion using Equation \ref{eqs:mappingSL2Ev} and then perform sampling:
\[
a^{(t)} \sim p \left( \mathbf{a} \vert \omega^{(t)} \right) = \mathtt{Cat}\left( \boldsymbol{\theta}^{(t)} \right) = \mathtt{Cat}\left( \theta^{(t)}_1, \theta^{(t)}_2 ..., \theta^{(t)} \right),
\]
where the parameters $\theta_i$ are computed as $\theta_i^{(t)} = b_i^{(t)} + u^{(t)} c_i$.

\subsection{Updating the opinion of actions} \label{ssec:Updating}

An opinion $\omega^{(t)}$ may be straightforwardly updated in the domain of evidential Dirichlet pdfs. Given the opinion $\omega^{(t)}$, we can recover the evidence in favor of each action by computing the evidence parameter $\mathbf{e}$ using Equation \ref{eq:tauW}:
\[
\mathbf{e}^{(t)}=\tau_{W}\left(\omega^{(t)}\right).
\]
If the observed reward $r^{(t)}$ support the belief that action $a^{(t)}$ is the best available, we can then update the evidence parameter simply adding one piece of evidence:
\[
\mathtt{(Generic Rule:)} \qquad e_i^{(t+1)}\leftarrow e_i^{(t)}+\eta \boldsymbol{1}\left[\textnormal{cdt}\right] \qquad \textnormal{for } a^{(t)}=a_{i}
\]
where $\eta\in \mathbb{R}_{>0}$ is a scalar controlling the size of the update, and $\boldsymbol{1}\left[\textnormal{cdt}\right]$ is the indicator function returning $1$ if the condition $\textnormal{cdt}$ is satisfied, otherwise returning zero. Thus, an update is executed on the evidence of the action just taken ($a^{(t)}=a_{i}$) if an update condition is met ($\textnormal{cdt}$).

We consider three possible conditions $\textnormal{cdt}$.
The first one states that evidence should be updated if the observed reward $r^{(t)}$ is greater than the mean of the observed average rewards $\hat{E} \left[ R_i^{(t)} \right]$:
\[
\mathtt{(Average Rule:)} \qquad e_i^{(t+1)}\leftarrow e_i^{(t)}+\eta \boldsymbol{1} \left[r^{(t)} > E_i\left[ \hat{E} \left[ R_i^{(t)} \right]  \right]\right] \qquad \textnormal{for } a^{(t)}=a_{i}.
\]
This rule may work well in initial stages, but it is bound to struggle to discriminate actions with close rewards. 
The second condition states that evidence should be updated if the observed reward $r^{(t)}$ is greater than the maximum of the estimated average rewards $\hat{E} \left[ R_i^{(t)}\right]$:
\[
\mathtt{(Max Rule:)} \qquad e_i^{(t+1)}\leftarrow e_i^{(t)}+\eta \boldsymbol{1} \left[r^{(t)} > \textnormal{max}_i\left[ \hat{E} \left[ R_i^{(t)} \right]  \right]\right] \qquad \textnormal{for } a^{(t)}=a_{i}.
\]
This rule is more stringent and it allows us to update only the actions that appear to be better than any other.
The third condition improves on the previous one, stating that the comparison should be with respect to the second-best action if we selected the current best action, otherwise with respect to the best action. Let us define $\hat{a}^{(t)*}$ and $\hat{a}^{(t)**}$ to be what we currently estimate as the best action and the second-to-best action, respectively, and let $\hat{E} \left[ R^{(t)*}\right]$ and $\hat{E} \left[ R^{(t)**}\right]$ the currently estimated expected reward associated with the best action and second-to-best action, respectively; then the new rule can be expressed as:
\[
\mathtt{(MaxRule2:)}\qquad\begin{cases}
e_{i}^{(t+1)}\leftarrow e_{i}^{(t)}+\eta\boldsymbol{1}\left[r^{(t)}>\hat{E}\left[R^{(t)**}\right]\right]\qquad\textrm{for }a^{(t)}=a_{i}=a^{(t)*}\\
e_{i}^{(t+1)}\leftarrow e_{i}^{(t)}+\eta\boldsymbol{1}\left[r^{(t)}>\hat{E}\left[R^{(t)*}\right]\right]\qquad\textrm{for }a^{(t)}=a_{i}\neq a^{(t)*}
\end{cases}
\] 

Finally, notice that the size of the update may be scaled not only statically, but also dynamically in proportion to the difference between the current reward $r^{(t)}$ and the expected reward $\hat{E} \left[ R_i^{(t)} \right]$ for the chosen action $i$ as:
\[ 
\eta = \zeta \left( r^{(t)} - \hat{E} \left[ R_i^{(t)} \right] \right),
\]
where $\zeta \in \mathbb{R}_{>0}$ is still a user-defined hyper-parameter.

Once the new evidence $\mathbf{e}^{(t+1)}$ has been computed, the corresponding opinion $\omega^{(t+1)}$ can then be recomputed using Equation \ref{eq:tauW-1}:
\[
\omega^{(t+1)}=\tau_{W,\mathbf{c}}^{-1}\left(\mathbf{e}^{(t+1)}\right),
\]
where $\mathbf{c}$ is the constant base rate of $\omega^{(t)}$.

\subsection{Uncertainty estimation}

As discussed in Section \ref{ssec:SL}, subjective logic provides us with simple and straightforward measures for the evaluation of uncertainty.
Given the agent opinion $\omega = \left( \boldsymbol{b}, u, \boldsymbol{c} \right)$, epistemic uncertainty may be assessed as the value of uncertainty $u$ in the opinion.
The overall uncertainty of the model, accounting for both aleatoric and epistemic uncertainty, can be computed as the entropy of the distribution defined by the opinion $\omega$ over the actions. A multinomial opinion $\omega$ induces a categorical distribution 
$
p\left(\mathbf{a}\vert\omega^{(t)}\right) = \mathtt{Cat} \left(
\boldsymbol{b}^{(t)} + u^{(t)} \boldsymbol{c}
\right).
$
From this distribution we can simply compute the overall uncertainty as the entropy of the distribution:
$
H[p]=-\sum_{i=1}^k p\left(a_i\vert \omega^{(t)}\right)\log p\left(a_i\vert \omega^{(t)}\right).
$
Although the two quantities (epistemic uncertainty and total uncertainty) are not directly comparable they will provide a way to gain an insight on how uncertainty evolves during learning.

\section{Empirical Evaluation \label{sec:EmpircalStudy}}

In this section we evaluate the SLB agent empirically from the point of view of performance and uncertainty estimation\footnote{Code for the simulations is available at \url{https://github.com/FMZennaro/SLBandits}}. Each multi-armed bandit problem is run for $1000$ episodes, and statistics are computed by averaging over $500$ trials.

\subsection{Simulation 1: Evaluating SLB on a standard testbed}
In this first simulation we evaluate the SLB algorithm assessing the performances of the variants of update rules we presented in Section \ref{ssec:Updating}, observing the effect hyper-parameters, and comparing its results against classical algorithms.

\paragraph{Setup.}
We consider a standard testbed $\mathcal{B}=\left\langle \mathcal{A},\mathcal{R}\right\rangle$ where the set $\mathcal{A} = \left\{ a_1, a_2, ..., \right.$ $\left. a_{10} \right\} $ contains ten actions, and where the set $\mathcal{R}= \left\{ \rho_1, \rho_2, ..., \rho_{10} \right\}$ contains ten Gaussian distribution $\rho_i = \texttt{Gauss} \left( \mu_i, \sigma_i\right)$ with $\mu_i \sim \texttt{Gauss} \left(0,1\right)$ sampled from a Normal distribution and $\sigma_i = 1$ \cite{sutton2018reinforcement}. 

We run four different SLB agents: an agent using the $\texttt{AverageRule}$ (\emph{SL(avg)}); an agent using the $\texttt{MaxRule}$ (\emph{SL(max)}); an agent using the $\texttt{MaxRule}$ with dynamic scaling (\emph{SL(maxs)}); and an agent using the $\texttt{MaxRule2}$ with dynamic scaling (\emph{SL(maxs2)}). All the agents have a single tunable hyper-parameter ($\eta$ or $\zeta$), and we explore different possible values for it in the set $\left\{ 0.1, 0.5, 1.0, 1.5, 2.0, 2.5,\right.$ $\left. 3.0, 5.0 \right\}$. For comparison, we run the following classical algorithms with the relative hyper-parameters tuned on a similar testbed of simulations in \cite{sutton2018reinforcement}: $\epsilon$-greedy agent with $\epsilon=0.1$, $\epsilon$-greedy agent with linear decay $\epsilon=\frac{1}{t^3}$, a UCB1-Normal agent with $c=2$ \cite{auer2002finite}, and a softmax gradient agent with $\eta_{gradient}=0.1$  \cite{sutton2018reinforcement}. We evaluate the performance as the percentage of times the agent selects the optimal action during each episode.

\begin{figure}
	\centering
	\centerline{\includegraphics[scale=.42]{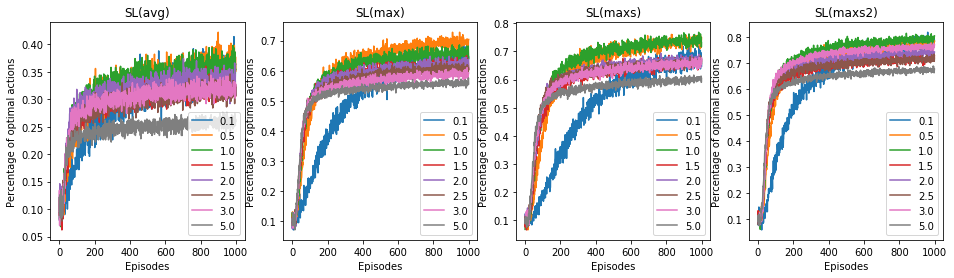}}	
	\caption{Percentage of optimal actions for the four SLB algorithms as a function of the episodes. \label{fig:sim1}}
\end{figure}

\paragraph{Results.}
Figure \ref{fig:sim1} shows the performance of the SLB algorithms. As expected, the $\texttt{MaxRule2}$ provides overall the best result (notice the different scale on the y-axis). For low values of the hyper-parameter, such as $\eta=0.1$ the agent converges very slowly towards an optimal policy; for high values, such as $\eta\geq 2.0$, the agent settles on a sub-optimal policy. On this testbed, all the variants of the SLB algorithm seem to achieve the best results for $\eta=0.5$ or $\eta=1.0$. 

\begin{figure}
	\centering
	\includegraphics[scale=.45]{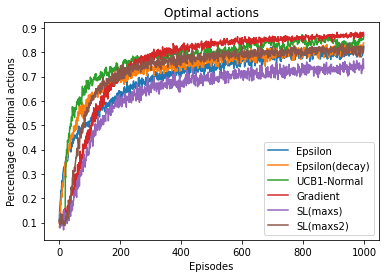}
	\caption{Percentage of optimal actions for SL(maxs), SL(maxs2), $\epsilon$-greedy, UCB, and gradient agents. \label{fig:sim2}}
\end{figure}

Selecting \emph{SL(maxs)} and \emph{SL(maxs2)} with $\eta=0.5$ as the best SLB candidates, we measure them against standard agents. Figure \ref{fig:sim2} shows that SL(maxs2) can reach a competitive performance on this testbed, reaching a percentage of correct actions higher than the $\epsilon$-greedy agent. Notice that this is remarkable considering that the SL(maxs2) always behaves probabilistically and never acts greedily, differently from the $\epsilon$-greedy or UCB agent that frequently select the best current action and differently from the gradient agent that uses a softmax or Boltzmann distribution to approximate a max operation. The lack of greedy exploitation is likely also be responsible for what looks like a sub-optimal behavior during the first episodes.

\subsection{Simulation 2: Assessing uncertainty in SLB}

A strength of the SLB algorithms follows from the possibility of evaluating its aleatoric and epistemic uncertainty, and in this simulation we examine the insights we may get from an analysis of uncertainty.  

\begin{figure}
	\centering
	\centerline{\includegraphics[scale=0.42]{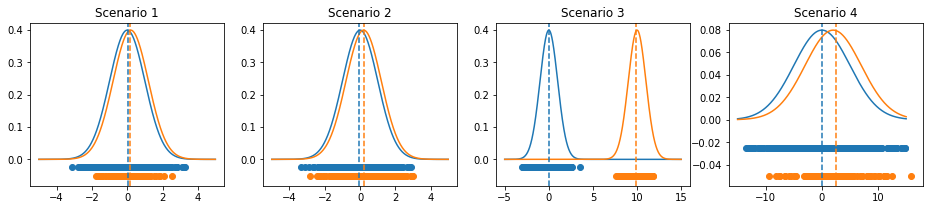}}
	\caption{Uncertainty scenarios. For each scenario, we reported the distribution of $\rho_1$ (in orange) and the distribution of $\rho_i$ (in blue); on the bottom we show $100$ samples from $\rho_1$ (in orange) and $100 \cdot (\left|\mathcal{A}\right|-1)$ samples from $\rho_i$ (in blue); we also plot the empirical average of this samples as a vertical dashed line. For more details on the settings of each scenario refer to the main text. \label{fig:graphs}}
\end{figure}

\paragraph{Setup.}
We consider four different scenarios in order to assess how the SLB algorithm behaves in terms of uncertainty estimation:
\begin{itemize}
	\item \emph{Scenario 1:} we consider $\mathcal{B}_1=\left\langle \mathcal{A},\mathcal{R}\right\rangle$ with $\left|\mathcal{A}\right| = 10$, a single reward pdf $\rho_1 = \texttt{Gauss}(0.2,1)$, and all the other ones $\rho_i = \texttt{Gauss}(0,1)$, for $2\leq i \leq 10$.
	
	\item \emph{Scenario 2:} we consider $\mathcal{B}_2=\left\langle \mathcal{A},\mathcal{R}\right\rangle$ with $\left|\mathcal{A}\right| = 2$, one reward pdf $\rho_1 = \texttt{Gauss}(0.2,1)$, and the other one $\rho_i = \texttt{Gauss}(0,1)$ for $i=2$.
	
	\item \emph{Scenario 3:} we consider $\mathcal{B}_3=\left\langle \mathcal{A},\mathcal{R}\right\rangle$ with $\left|\mathcal{A}\right| = 10$, a single reward pdf $\rho_1 = \texttt{Gauss}(10,1)$, and all the other ones $\rho_i = \texttt{Gauss}(0,1)$, for $2\leq i \leq 10$.
	
	\item \emph{Scenario 4:} we consider $\mathcal{B}_4=\left\langle \mathcal{A},\mathcal{R}\right\rangle$ with $\left|\mathcal{A}\right| = 10$, a single reward pdf is $\rho_1 = \texttt{Gauss}(2,5)$, and all the other ones $\rho_i = \texttt{Gauss}(0,5)$, for $2\leq i \leq 10$.
\end{itemize}
Figure \ref{fig:graphs} offers a graphical illustration of how these scenarios provide different degrees of challenge in concluding that action $a_1$ (in orange) is superior to actions $a_i$ (in blue). In Scenario 1 and 2 the gap between the means is extremely narrow; in Scenario 4 the gap is wider but the pdfs have higher variance; in contrast, Scenario 3 represents an easy problem. Notice, also, that while these images display a static set of samples for each action, the sampling when running a multi-armed bandit is dynamic and dependent on the current estimates; Scenario 1, with its higher number of actions, is therefore more challenging than Scenario 2, with only two actions.  
We run the \emph{SL(maxs2)} agent and we analyze how its epistemic and total uncertainty evolve during the episodes.

\begin{figure}
	\centering
	\includegraphics[scale=0.45]{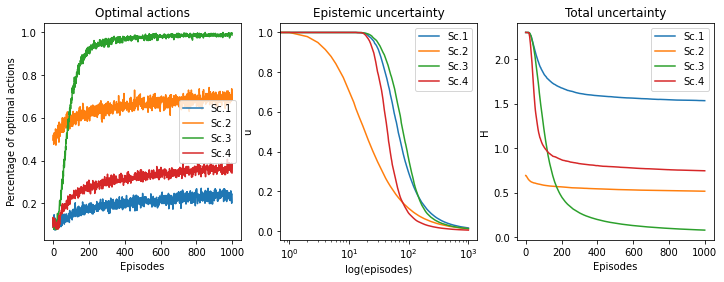}
	\caption{Percentage of optimal actions, epistemic uncertainty and total uncertainty for the agent SL(maxs) across the four scenarios. Notice the logarithmic scale in the second graph. \label{fig:sim4a}}
\end{figure}

\paragraph{Results.}
Figure \ref{fig:sim4a} shows the performance of the SLB agent (percentage of optimal actions), its epistemic uncertainty (parameter $u^{(t)}$ of the current opinion $\omega^{(t)}$), and its total uncertainty (entropy of the categorical distribution over the actions implied by the current opinion $\omega^{(t)}$).

First of all, the overall performance follows the ranking of difficulty we discussed above. Scenario 3 is easily solved achieving a perfect performance, but, on the other scenarios, the SLB agent struggles to identify the correct option and the performance is marginally above random guess.
Second, there is a correlation between the evolution of the performance and the uncertainties: they change mostly within the first $200$ episodes and then they enter in a saturation regime. Learning at this point is slowed down by the amount of evidence accumulated. Uncertainty may then be used as a proxy to assess the learning saturation of the agent in this model.
Third, in the short-term regime, we can observe different dynamics in the drop of epistemic uncertainties. 
The faster drop in Scenario 2 can be explained by the restricted number of actions: at each episode, the collected reward $r^{(t)}$ has to be compared against the expected rewards $\hat{E} \left[ R_i^{(t)} \right]$ of only two actions, and therefore it is more likely that the update condition is met. This steep reduction in uncertainty reflects the fact that we collect informative samples more frequently. 
The fast drop in Scenario 4 is due, instead, to the high variance of the setting: in each episode, there is more chance of drawing a high-value reward $r^{(t)}$ that is higher than the current expected rewards $\hat{E} \left[ R_i^{(t)} \right]$ of the other actions. Differently from the result on Scenario 2, these dynamics may be undesired, as here our epistemic uncertainty decreases because of possible outliers.  
Fourth, in the long term-regime, we can infer something about aleatoric uncertainty. Although epistemic uncertainty and total uncertainty are incommensurable as they are measured in different units, we can still observe their combined evolution.
Remarkably, and expectedly, the total uncertainty in Scenario 3 drops to a very low level. Thus, given that its epistemic uncertainty is relatively high, its aleatoric uncertainty must be very low. This makes sense, as in this scenario there is little room for uncertainty in the intrinsic dynamics of the system regarding which action is the best.
In Scenario 2, the total uncertainty flattens very early, meaning that despite the accumulation of more evidence signaled by the drop in epistemic uncertainty, the aleatoric uncertainty does not decrease significantly; the agent is stuck on a distribution that is not a uniform distribution (1 bit), but it is far from having determined an absolute best action either.

These insights on the evolution of uncertainty allows us to draw a more informed conclusion about the result of the training process, and thus orient our action. The dynamics of epistemic uncertainty may provide a hint to decide up to what points collecting new samples may provide useful information for learning; the evolution of aleatoric uncertainty may offer information on when the process of training is slowed down by the barrier of the intrinsic randomness, and what the magnitude of this randomness may be.

\section{Discussion and Future Work \label{sec:Discussion}} 

The SLB algorithm offers a way to integrate and exploit the interpretability of subjective logic in tackling the multi-armed bandit problem. Our approach exploits the duality in the representation of subjective opinions, as beliefs and Dirichlet pdfs.  Although it does not operate completely in the subjective logic domain (sampling works with standard probability distributions, updating works with evidential Dirichlet pdfs), multinomial opinions allows for an intuitive and immediate assessment of the evolution of uncertainties.
The algorithms we presented are only a possible way to integrate subjective logic in the bandit problem, and in the future we plan to explore improvements of the current algorithms. More importantly, a careful theoretical analysis of the regret of the SLB agent and the dynamics of uncertainty, especially in comparison with other standard and Bayesian algorithms, is part of our ongoing research.

\section{Conclusion  \label{sec:Conclusion}}

The multi-armed bandit problem constitutes a simple learning problem that may be seen as a basic element of more complex problems (contextual bandits, non-stationary bandits, full reinforcement learning). Understanding and characterizing the dynamics of uncertainty in this simplified setting will allow for extensions to more realistic settings.
In this work we analyzed the possibility of using subjective logic to estimate the evolution of uncertainty in multi-armed bandit problems. We formulated different variants of agents based on subjective logic, we evaluated their performance, and showed that they can perform competitively against standard algorithms; moreover, we demonstrated that our agents allow for an estimation of aleatoric and epistemic uncertainty, and that this information may be used to obtain insights into the learning process of the agent. Although preliminary, our results suggest that the use of subjective logic offers useful insights that may be exploited by more sophisticated agents.

%
%
%
\bibliographystyle{splncs04}
\bibliography{../../lib/lib}
%
%
%
%
%
\end{document}